\documentclass[runningheads]{llncs}
\usepackage{graphicx}
\usepackage{epstopdf}
\usepackage{times}
\usepackage{soul}
\usepackage{url}
\usepackage[hidelinks]{hyperref}
\usepackage[utf8]{inputenc}
\usepackage[small]{caption}
\usepackage{graphicx}
\usepackage{amsmath}
\usepackage{booktabs}
\usepackage{algorithm}
\usepackage{algorithmic}
\usepackage[marginal]{footmisc}

\usepackage{amssymb} 
\usepackage{makecell}

\begin{document}
\title{Virtual Adversarial Training on Graph Convolutional Networks in Node Classification}

\author{Ke Sun\inst{1} \and
Zhouchen Lin\inst{2} \and
Hantao Guo\inst{1} \and
Zhanxing Zhu\inst{3,1,4}
}

%
%
\institute{Center for Data Science, Peking University, China \and
Key Laboratory of Machine Perception (MOE), School of EECS, Peking University, China \and
School of Mathematical Science, Peking University, China \and
Beijing Institute of Big Data Research (BIBDR), China
\\
\email{\{ajksunke,zlin,guohantao,zhanxing.zhu\}@pku.edu.cn}}
\maketitle              
\begin{abstract}
The effectiveness of Graph Convolutional Networks~(GCNs) has been demonstrated in a wide range of graph-based machine learning tasks. However, the update of parameters in GCNs is only from labeled nodes, lacking the utilization of unlabeled data. In this paper, we apply Virtual Adversarial Training~(VAT), an adversarial regularization method based on both labeled and unlabeled data, on the supervised loss of GCN to enhance its generalization performance. By imposing virtually adversarial smoothness on the posterior distribution in semi-supervised learning, VAT yields an improvement on the performance of GCNs. In addition, due to the difference of property in features, we perturb virtual adversarial perturbations on sparse and dense features, resulting in GCN Sparse VAT~(GCNSVAT) and GCN Dense VAT~(GCNDVAT) algorithms, respectively. Extensive experiments verify the effectiveness of our two methods across different training sizes. Our work paves the way towards better understanding the direction of improvement on GCNs in the future.
\end{abstract}

\section{Introduction}
Recently, research of analyzing graphs with machine learning has received more and more attention, mainly focusing on node classification~\cite{kipf2016semi}, link prediction~\cite{zhu2016max} and clustering tasks~\cite{fortunato2010community}. 	Graph convolutions, as the transformation of traditional convolutions from Euclidean domain to non-Euclidean domain, have been leveraged to design Graph Neural Networks to deal with a wide range of graph-based machine learning tasks. 

Graph Convolutional Networks~(GCNs)~\cite{kipf2016semi} generalize convolutional neural networks~(CNNs) to graph structured data from the perspective of spectral theory based on prior works~\cite{bruna2013spectral,defferrard2016convolutional}. It has been demonstrated that GCN and its variants~\cite{hamilton2017inductive,velickovic2017graph} significantly outperform traditional multi-layer perceptron~(MLP) models and prior graph embedding approaches~\cite{tang2015line,perozzi2014deepwalk,grover2016node2vec}. 

\footnote{\noindent This is one oral paper published in PRCV 2019 and the first author is a student.}

However, there are still many deficits on GCNs, thus in this paper we propose to apply VAT on GCNs to tackle these drawbacks of GCNs. Particularly, we firstly highlight the importance of VAT on GCNs from the following aspects, which construct the motivation of our approaches.

\noindent \textbf{Lacking the Leverage of Unlabeled Data for GCNs.} The optimization of GCNs is solely based on the labeled nodes. Concretely speaking, GCNs directly distribute gradient information over the entire labeled set of nodes from the supervised loss. Due to the lack of loss on unlabeled data, the parameters that are not involved in the receptive field may not be updated~\cite{chen2017stochastic}, resulting in the inefficiency of information propagation of GCNs. 

\noindent \textbf{Effect of Regularization in Semi-Supervised Learning.} Regularization plays a crucial role in semi-supervised learning including graph-based learning tasks. On the one hand, by introducing regularization, a model can make full use of unlabeled data, thus enhancing the performance in semi-supervised learning. On the other hand, regularization can also be regarded as prior knowledge that can smooth the posterior output. For GCN model, a good regularization can not only leverage the unlabeled data to refine its optimization, but only benefit the performance of GCNs, resulting in a improved generalization performance.

\noindent \textbf{Virtual Adversarial Regularization on GCNs.} Virtual Adversarial Training~(VAT)~\cite{miyato2018virtual} smartly performs adversarial training without label information to impose a local smoothness on the classifier, which is especially beneficial to semi-supervised learning. In particular, VAT endeavors to smooth the model anisotropically in the direction in which the model is the most sensitive, i.e., the adversarial direction, to improve the generalization performance of a model. In addition, the existence of robustness issue in GCNs has been explored in recent works~\cite{zugner2018adversarial,zugner2018adversarial2}, allowing VAT on graph-based learning task. 

Due to the fact that VAT has been successfully applied on semi-supervised image classification~\cite{miyato2018virtual,yu2018tangent} and text classification~\cite{miyato2016adversarial}, a natural question could be asked: \textit{Can we utilize the efficacy of VAT to improve the performance of GCNs in semi-supervised node classification?}

Following this motivation, in our paper, we formally introduce VAT regularization on the original supervised loss of GCNs in semi-supervised node classification task. Concretely speaking, firstly, a detailed analysis of GCNs focusing on the first-order approximation of local spectral convolutions is provided to demystify how GCNs work in semi-supervised learning. Moreover, based on the motivation described above, we elaborate the process of applying VAT on GCNs in a theoretical way by additionally imposing virtual adversarial loss on the basic loss of GCNs, resulting in GCNVAT algorithm framework. Next, due to the sparse property of node features, in the realization of our method, we actually add virtual adversarial perturbations on sparse and dense features, respectively, and attain the GCNSVAT and GCNDVAT algorithms. Finally, in the experimental part, we demonstrate the effectiveness of the two approaches under different training sizes and refine a theoretical analysis on the sensitivity to the hyper-parameters on VAT, facilitating us to apply our approaches in real applications involving graph-based machine learning tasks. In summary, the contributions of the paper are listed below:

\begin{itemize}
	\item To the best of our knowledge, we are the first to focus on applying better regularization on original GCN to refine its generalization performance.
	\item We are the first to successfully transfer the efficacy of Virtual Adversarial Training~(VAT) to the semi-supervised node classification on graphs and point out the difference compared with image and text classification setting.
	\item We refine the sensitivity analysis of hyper-parameters in GCNSVAT and GCNDVAT algorithms, facilitating the deployment of our methods in real scenarios.
\end{itemize}

\section{GCNs with Virtual Adversarial Training}
In this section, we will elaborate how GCNs work in semi-supervised learning and how to utilize the virtual adversarial training to smooth the posterior distribution of GCNs.
\subsection{Semi-Supervised Classification with GCNs}
Firstly, we denote a graph by $G=(V,E)$, where $V$ is the vertex set and $E$ is the edge set. $X$ and $A$ are the features and adjacent matrix of the graph, respectively and $D=\text{diag}(d_1,d_2,\cdots,d_n)$ denotes the degree matrix of $A$, where $d_i=\sum_{j}a_{ij}$ is the degree of vertex $i$.

\noindent \textbf{First-Order Approximation.} GCNs are based on the graph spectral theory. For efficient computation, \cite{defferrard2016convolutional} approximate the spectral filter $g_{\theta}$ with Chebyshev polynomials up to $K^{th}$ order:
\begin{equation} \begin{aligned} 
g_{\theta^\prime}(\Lambda)=  \sum_{k=0}^{K-1} {\theta_k^\prime T_k(\Lambda) },
\end{aligned} \end{equation}
where $\Lambda$ is the eigenvalues matrix of normalized graph Laplacian $L=I_N-D^{-\frac{1}{2}}AD^{-\frac{1}{2}}$. $T_k$ is the Chebyshev polynomials and $\theta_k^\prime$ is a vector of Chebyshev coefficients. Further, \cite{kipf2016semi} simplified this model by limiting $K=1$ and then the first-order approximation of spectral graph convolution is defined as:
\begin{equation} \begin{aligned} 
g_\theta \star x = \theta(I_N+D^{-\frac{1}{2}}AD^{-\frac{1}{2}}) x,
\end{aligned} \end{equation}
where $\theta$ is the only Chebyshev coefficients left. Through the normalization trick, the final form of graph convolutional networks with two layers in GCNs~\cite{kipf2016semi} is:
\begin{equation} \begin{aligned} 
Z=f(X,A)=\text{softmax}(\hat{A} \ \text{ReLU}(\hat{A}XW^{(0)}) W^{(1)}),
\end{aligned} \end{equation}
where $\hat{A}=\tilde{D}^{-\frac{1}{2}}\tilde{A}\tilde{D}^{-\frac{1}{2}}, \tilde{A}=A+I$. $\tilde{D}$ is the degree matrix of $\tilde{A}$. $Z$ is the obtained embedding matrix from nodes,  $W^{(0)}$ is the input-to-hidden weight matrix and $W^{(1)}$ is the hidden-to-output weight matrix.

\noindent \textbf{Optimization.} Finally, the loss function is defined as the cross entropy error over all labeled nodes:
\begin{equation} \begin{aligned} 
\mathcal{L}_0 = -\sum_{l \in \mathcal{Y}_L} \sum_{f=1}^{F} Y_{lf} Z_{lf},
\end{aligned} \end{equation}
where $\mathcal{Y}_L$ is the set of node indices that have labels. In fact, the performance of GCNs heavily depends on the efficiency of this Laplacian Smoothing Convolutions, which has been demonstrated in \cite{li2018deeper,kipf2016semi}. Therefore, how to design a good regularization to smooth the posterior distribution of GCNs plays a crucial role for the improvement of performance for GCNs.

\subsection{Virtual Adversarial Training in GCNs}
Virtual Adversarial Training~(VAT)~\cite{miyato2018virtual} is a regularization method that trains the output distribution to be isotropically smooth around each input data point by selectively smoothing the model in its most anisotropic direction, namely adversarial direction. In this section, we apply VAT on GCNs to smooth the posterior distribution of GCNs. 

\noindent \textbf{Assumptions.} Firstly, both VAT and GCNs mainly focus on semi-supervised setting, in which two assumptions should be implicitly met~\cite{yu2018tangent}:
\begin{itemize}
	\item \textbf{Manifold Assumption.} The observed data $x$ presented in high dimensional space is with high probability concentrated in the vicinity of some underlying manifold with much lower dimensional space.
	\item \textbf{Smoothness Assumption.} If two points $x_1, x_2 \in \mathcal{M}$ are close in manifold distance, then the conditional probability $p(y|x_1)$ and $p(y|x_2)$ should be similar. In other words, the true classifier, or the true condition distribution $p(y|x)$ varies smoothly along the underlying manifold $\mathcal{M}$.
\end{itemize}
In the node classification task, GCNs, which involve the graph embedding process, also implicitly conform to these assumptions. There is underlying manifold in the process of graph embedding and the conditional distribution of embedding vectors are expected to vary smoothly along the underlying manifold. In this way, we are capable of utilizing VAT to smooth the embedding of nodes in the adversarial direction to improve the generalization of GCNs. 

\noindent \textbf{Difference of VAT on Graph and Image, Text.} Traditional VAT~\cite{miyato2018virtual} is proposed on image classification while VAT on text classification~\cite{miyato2016adversarial} is applied on word embedding vectors of each word. For VAT on graphs, we simply apply VAT on the features of nodes for easy implementation. Additionally, another obvious difference lies in that the relation between each node is not independent for the node classification task compared with image and text classification. The classification result of each node not only depends on the feature itself but also the features of its neighbors, resulting in the \textit{Propagation Effect} of perturbations on feature of each node. We use $\mathcal{D}_l$ and $\mathcal{D}_{ul}$ to denote dataset with labeled nodes and unlabeled nodes respectively. $\overline{x}$ represents features excluding feature $x$ of current node.

\noindent \textbf{Adversarial Training in GCNs.} Here we formally define the adversarial training in GCNs, where adversarial perturbations are solely added on features of labeled nodes:
\begin{equation} \begin{aligned} 
\min_{\theta} \max_{r,\|r\|\leq\epsilon} D\left[q(y|x_l, \overline{x}, A), p(y|x_l+r, \overline{x}, A;\theta)\right],
\end{aligned} \end{equation}
where $D[q,p]$ measures the divergence between two distributions $q$ and $p$. $q(y|x_l, \overline{X}, A)$ is the true distribution of output labels, usually one hot vector $h(y;y_l)$ and $p(y|x_l+r, \overline{x}, A)=f(X,A)$ denotes the predicted distribution by GCNs. $x_l$ represents the feature of current labeled node and $r$ represents the adversarial perturbation on the feature $x_l$. When the true distribution is denoted by one hot vector $h(y;y_l)$, the perturbation $r_{\rm adv}$ in $L_{2}$ norm can be linearly approximated: 
\begin{equation} \begin{aligned} 
r_{\rm adv} \approx  \epsilon \frac{g}{\|g\|_2},\ {\rm where}\ g=\nabla_{x_l} D\left[h(y;y_l), p(y|x_l, \overline{x}, A;\theta)\right].
\end{aligned} \end{equation}

\noindent \textbf{Virtual Adversarial Loss.} In order to utilize the unlabeled data, we are expected to evaluate the true conditional probability $q(y|x_l, \overline{x}, A)$. Therefore, we use the current estimate $p(y|x,\overline{x}, A;\hat{\theta})$ in place of $q(y|x, \overline{x}, A)$.
\begin{equation} \begin{aligned} 
\min_{\theta} \max_{r,\|r\|\leq\epsilon} D\left[p(y|x,\overline{x}, A;\hat{\theta}), p(y|x+r,\overline{x}, A;\theta)\right]
\end{aligned} \end{equation}
Then virtual adversarial regularization is constructed from inner max loss:
\begin{equation}\begin{aligned} 
\mathcal{R}_{\rm vadv}(x, \mathcal{D}_l, \mathcal{D}_{ul}, \theta) =  \max_{r,\|r\|\leq\epsilon} D\left[p(y|x,\overline{x}, A;\hat{\theta}), p(y|x+r,\overline{x}, A;\theta)\right]
\end{aligned} \end{equation}
The final regularization term we propose in this
study is the average of $\mathcal{R}_{\rm vadv}(x, \mathcal{D}_l, \mathcal{D}_{ul}, \theta)$ over all input nodes:
\begin{equation}
\mathbb{E}_{x\sim \mathcal{D}}\mathcal{R}_{\rm vadv}=\frac{1}{N_l + N_{ul}}\sum_{x \in \mathcal{D}_l, \mathcal{D}_{ul}} 	\mathcal{R}_{\rm vadv}(x, \mathcal{D}_l, \mathcal{D}_{ul}, \theta) 
\end{equation}
\noindent \textbf{Virtual Adversarial Training.}
The full objective function is thus given by:
\begin{equation}
\min_{\theta} \mathcal{L}_0 +  \alpha\mathbb{E}_{x\sim \mathcal{D}}\mathcal{R}_{\rm vadv}, 
\end{equation}
where $\mathcal{L}_0$ is constructed from labeled nodes in GCNs, $\alpha$ denotes the regularization coefficient and VAT regularization is crafted from both labeled and unlabeled nodes. 

\subsection{Fast Approximation of VAT in GCNs}

The key of VAT in GCNs is the approximation of $r_{\text{vadv}}$ where
\begin{equation} \begin{aligned} 
r_{\rm vadv} = \mathop{\arg\max}_{r,\|r\|\leq\epsilon} D\left[p(y|x,\overline{x}, A;\hat{\theta}), p(y|x+r,\overline{x}, A;\hat{\theta})\right].
\end{aligned} \end{equation}

\noindent \textbf{Second-Order Approximation.} Just like the situation in traditional VAT, the evaluation of GCNs with VAT cannot be performed with the linear approximation since first-order approximation equals zero~\cite{miyato2016adversarial}. Therefore, a second-order approximation is needed:
\begin{equation} \begin{aligned} 
D(r,x,\overline{x}, A;\hat{\theta}) \approx \frac{1}{2}r^{T}H(x,\overline{x}, A; \hat{\theta}) r,
\end{aligned} \end{equation}
where $ H(x,\overline{x}, A; \hat{\theta}) := \nabla_{r}^2 D(r,x,\overline{x}, A;\hat{\theta})|_{r=0}$. Then the evaluation of $r_{\rm vadv}$ can be approximated by:
\begin{equation} \begin{aligned} 
r_{\rm vadv} \approx \arg \mathop {\rm max}\limits_r \{r^{T} H(x, \hat{\theta}) r; ~\|r\|_2 \leq \epsilon\} = \epsilon \overline{u(x, \overline{x}, A;\hat{\theta})}, 
\end{aligned} \end{equation}
where $u(x,\overline{x}, A;\hat{\theta})$ is the first dominant eigenvector  of $H(x,\overline{x}, A;\hat{\theta})$ with magnitude 1.

\noindent \textbf{Power Iteration and Finite Difference Approximation.} After power interation and finite difference approximation mentioned in \cite{miyato2016adversarial}, the final approximation of $r_{\rm vadv}$ is:
\begin{equation} \begin{aligned} 
r_{\rm vadv} \approx  \epsilon \frac{g}{\|g\|_2}, {\rm where}\ g=\nabla_r D\left[p(y|x,\overline{x}, A;\hat{\theta}), p(y|x+r,\overline{x}, A; \hat{\theta})\right]\Big|_{r=\xi d}.
\end{aligned} \end{equation}

\section{Algorithm}

In this section, we will elaborate our Graph Convolutional Networks with Virtual Adversarial Training (GCNVAT) Algorithm. Algorithm~\ref{alg:GCNVAT} summarizes the procedures of the computation of mini-batch SGD for GCNs with VAT algorithm. 

\begin{algorithm}[ht]
	\caption{Mini-batch SGD for GCNVAT Framework}
	\label{alg:GCNVAT}
	\textbf{Input}: Features Matrix $X$, Adjacent Matrix $A$. Graph Convolution Network $f_{\theta}$\\
	\textbf{Output}: Graph Embedding $Z=f_{\theta}(X,A)$  
	
	\begin{algorithmic}[1] 
		\STATE Choose $M$ samples of $x^{(i)} (i=1,\dots,M)$ from dataset $\mathcal{D}$ at random.
		\STATE Compute the predicted distribution of current GCNs:
		$$p(y|x_l, \overline{x}, A;\hat{\theta}) \leftarrow f_{\hat{\theta}}(X,A)$$
		
		\STATE \textbf{\% Step 1: Fast Approximation of $r_{\rm vadv}$}	
		\STATE Generate a random unit vector $d^{(i)} \in R^{I}$ using an iid Gaussian distribution.
		\STATE Calculate $r_{\rm vadv}$ via taking the gradient of $D$ with respect to $r$ on $r=\xi d^{(i)}$ on each input data point $x^{(i)}$:
		\begin{equation} \begin{aligned}
		g^{(i)} \leftarrow 
		\nabla_{r} D\left[p(y|x^{(i)},\overline{x}, A; \hat{\theta}), p(y|x^{(i)}+r,\overline{x}, A;\hat{\theta})\right]\Big|_{r=\xi d^{(i)}} \nonumber
		\end{aligned} \end{equation}
		
		\STATE Evaluation of $r_{\rm vadv}$:
		$$r_{\rm vadv}^{(i)} \leftarrow \epsilon g^{(i)}/\|g^{(i)}\|_2$$
		
		\STATE \textbf{\% Step 2: Evaluation of Virtual Adversarial Loss}
		\begin{equation} \begin{aligned}
		\mathbb{E}_{x\sim \mathcal{D}}\mathcal{R}_{\rm vadv} = 
		\nabla_{\theta} \left(\frac{1}{M}\sum_{i=1}^M D\left[p(y|x^{(i)}; \hat{\theta}), p(y|x^{(i)}+r_{{\rm vadv}}^{(i)}; \theta)\right]\right)\Bigg|_{\theta=\hat{\theta}} \nonumber
		\end{aligned} \end{equation}
		
		\STATE \textbf{\% Step 3: Virtual Adversarial Training}
		\STATE Compute the supervised loss $\mathcal{L}_0$ of GCNs:
		\begin{equation} \begin{aligned} 
		\mathcal{L}_0 = -\sum_{l \in \mathcal{Y}_L} \sum_{f=1}^{F} Y_{lf} Z_{lf} \nonumber
		\end{aligned} \end{equation}
		
		\STATE Update $\theta$ by optimizing the full objective function $\mathcal{L}$:
		\begin{equation}
		\mathcal{L}=\mathcal{L}_0 +  \alpha\mathbb{E}_{x\sim \mathcal{D}}\mathcal{R}_{\rm vadv} \nonumber
		\end{equation}
	\end{algorithmic}
\end{algorithm}

Our GCNVAT Algorithm Framework is economical in computation since the derivative of the full objective function can be computed with at most three sets of propagation in total. Specifically speaking, firstly, by initializing the random unit vector $d^{(i)}$ in mini-batch and computing the gradient of divergence between predicted distribution of GCNs and that with the initial perturbation, we can evaluate the fast approximated $r_{\rm vadv}$, which is involved in the first set of back propagation. Secondly, after the computation of $r_{\rm vadv}$, we are able to compute the average virtual adversarial loss in the mini-batch and optimize this loss under fixed $r_{\rm vadv}$, which incorporates the second set of back propagation. Finally, the third back propagation is related to the original supervised loss based on labeled nodes in GCNs. All in all, by this GCNs with VAT algorithm including three sets of back propagation, we are capable of imposing the local adversarial regularization on the original supervised loss of GCNs through smoothing the posterior distribution of the model in the most adversarial direction, thereby improving the generalization of original GCNs.

\noindent \textbf{GCNSVAT and GCNDVAT} In the real scenarios, there are usually sparse features for each node especially for a large graph, which are involved in the computation of sparse tensor. In this case, in the implementation of our GCNVAT algorithm framework, we customize two similar GCNVAT methods for  different properties of node features. For GCN Sparse VAT~(GCNSVAT), we only apply virtual adversarial perturbations on the specific sparse elements in feature of each node, which may save much computation time especially for high-dimensional feature vectors. For GCN Dense VAT~(GCNDVAT), we actually perturb each element in feature by transforming the the sparse feature matrix to a dense one.

\section{Experiments}

In the experimental part, we conduct extensive experiments to demonstrate the effectiveness of our GCNSVAT and GCNDVAT algorithms. Firstly, we test the performance of both algorithms under different label rates compared with the original GCN. Then we make another comparison under the standard semi-supervised setting with other state-of-the-art approaches. Finally, a sensitivity analysis of hyper-parameters is provided for broad deployment of our method in real applications.

\noindent \textbf{Experimental Setup} For the graph dataset, we select the three commonly used citation networks: CiteSeer, Cora and PubMed~\cite{sen2008collective}. Dateset statistics are summarized in Table~\ref{table_dataset}. For all methods involved in GCNs, we use the same hyper-parameters as in \cite{kipf2016semi}: learning
rate of 0.01, 0.5 dropout rate, 2 convolutional layers, and 16 hidden units without
validation set for fair comparison. As for the hyper-parameters, we fix regularization coefficient $\alpha=1.0$ and only change the perturbation magnitude $\epsilon$ to control the regularization effect under different training sizes, which is further discussed later in the sensitivity analysis part. All the results are the mean accuracy of 10 runs to avoid stochastic effect.

\begin{table}[htbp]
	\centering
	\begin{tabular}{lrrrrr}
		\hline
		\textbf{Dateset}  & \textbf{Nodes} & \textbf{Edges} & \textbf{Classes} & \textbf{Features} & \textbf{Label Rate}\\
		\hline
		CiteSeer &    3327&    4732&    6&    3703&  3.6\%\\
		Cora     &    2708&    5429&    7&    1433&  5.2\%\\
		PubMed   &   19717&   44338&    3&     500&  0.3\%\\
		\hline
	\end{tabular}
	\caption{Dateset statistics}
	\label{table_dataset}
\end{table}

\subsection{Effect under Different Training Sizes}

\begin{figure}[htbp]
	\centering
	\centering\includegraphics[width=0.8\textwidth,trim=100 0 80 50,clip]{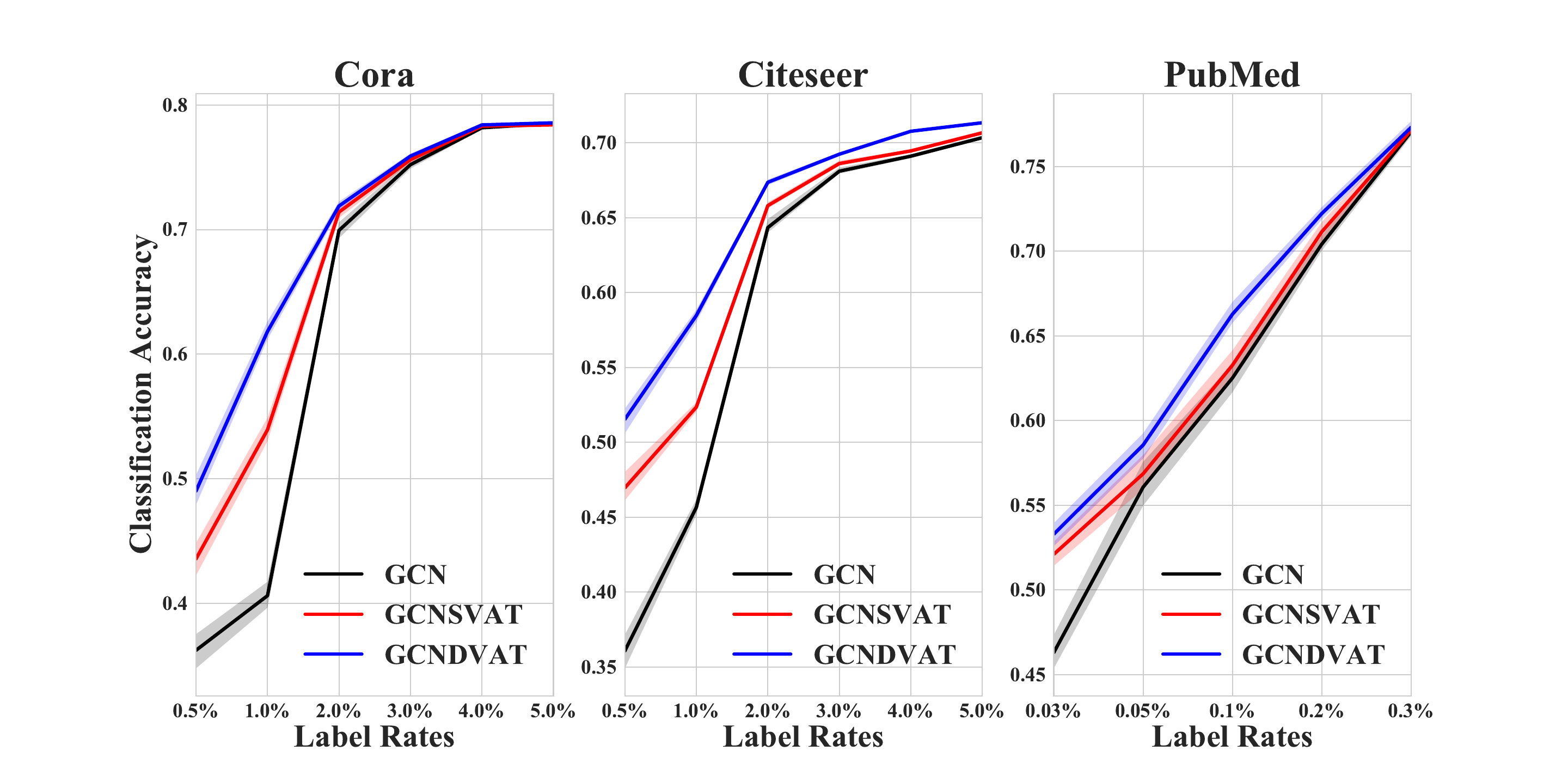}
	\caption{Classification Accuracies of GCNSVAT and GCNDVAT algorithms compared with GCN across three datasets.}
	\label{labelrate}
\end{figure}

To verify the consistent effectiveness of our two methods on the improvement of generalization performance, we compare GCNSVAT and GCNDVAT algorithms with original GCN method~\cite{kipf2016semi} under different training sizes across the three datasets and the results can be observed in Figure~\ref{labelrate}.

As illustrated in Figure~\ref{labelrate}, GCNSVAT~(the red line) and GCNDVAT~(the blue line) outperform original GCN~(the black line) consistently under all tested label rates. Actually, it is important to note that with the increasing of label rates, the regularization effect imposed by VAT on GCNs diminishes in both approaches since the improvement from regularization based on unlabeled data is decreasing. In other words, the superior performance of GCN with Virtual Adversarial Training are especially significant when there are few training sizes. Fortunately, in real scenarios, it is common to observe graphs with a small number of labeled nodes, thereby our algorithms are especially practical in these applications.

\noindent \textbf{Choice of GCNSVAT and GCNDVAT.} GCNDVAT performs consistently better in comparison with GCNSVAT even though GCNDVAT requires extra computation cost related to perturbations in the entire feature space. As for the reason, we argue that continuous perturbations in features facilitate the effect of VAT than discrete perturbations in sparse features. However, in the scenarios where the graph are large-scaled and their features are sparse, it is more appropriate to utilize GCNSVAT from the perspective of economical computation.

\begin{table}[ht]
	\centering\textbf{Cora}
	\begin{tabular}{lrrrrrr}
		\hline
		\textbf{Rates}  & 0.5\% & 1\% & 2\% & 3\% & 4\% & 5\%\\
		\hline
		\makecell{\textbf{GCN}}&\makecell{36.2\\(0.11)}&\makecell{40.6\\(0.08)}&\makecell{69.0\\(0.05)}&\makecell{75.2\\(0.03)}&\makecell{78.2\\(0.1)}&\makecell{78.4\\(0.01)}\\
		\hline
		\makecell{\textbf{GCN}\\\textbf{SVAT}}&\makecell{43.6\\(0.10)}&\makecell{53.9\\(0.08)}&\makecell{71.4\\(0.05)}&\makecell{75.6\\(0.02)}&\makecell{78.3\\(0.01)}&\makecell{78.5\\(0.01)}\\
		\hline
		\makecell{\textbf{GCN}\\\textbf{DVAT}}&\makecell{\bf49.0\\(0.10)}&\makecell{\bf61.8\\(0.06)}&\makecell{\bf71.9\\(0.03)}&\makecell{\bf75.9\\(0.02)}&\makecell{\bf78.4\\(0.01)}&\makecell{\bf78.6\\(0.01)}\\
		\hline
	\end{tabular}
	\caption{Classification Accuracies on Cora with different label rates. Numbers in bracket are the standard deviation of accuracies.}
	\label{table_weakly_cora}
\end{table}
\begin{table}[ht]	
	\centering{\textbf{CiteSeer}}
	\begin{tabular}{lrrrrrr}
		\hline
		\textbf{Rates}  & 0.5\% & 1\% & 2\% & 3\% & 4\% & 5\%\\
		\hline
		\makecell{\textbf{GCN}}&\makecell{36.1\\(0.09)}&\makecell{45.7\\(0.04)}&\makecell{64.3\\(0.04)}&\makecell{68.1\\(0.02)}&\makecell{69.1\\(0.01)}&\makecell{70.3\\(0.01)}    \\
		\hline
		\makecell{\textbf{GCN}\\\textbf{SVAT}}&\makecell{47.0\\(0.08)}&\makecell{52.4\\(0.02)}&\makecell{65.8\\(0.02)}&\makecell{68.6\\(0.01)}&\makecell{69.5\\(0.01)}&\makecell{70.7\\(0.01)}    \\
		\hline
		\makecell{\textbf{GCN}\\\textbf{DVAT}}&\makecell{\bf51.5\\(0.07)}&\makecell{\bf58.5\\(0.03)}&\makecell{\bf67.4\\(0.01)}&\makecell{\bf69.2\\(0.01)}&\makecell{\bf70.8\\(0.01)}&\makecell{\bf71.3\\(0.01)}\\
		\hline
	\end{tabular}
	\caption{Classification Accuracies on CiteSeer with different label rates. Numbers in bracket are the standard deviation of accuracies.}
	\label{table_weakly_citeseer}
\end{table}
\begin{table}[htbp]	
	\centering\textbf{PubMed}
	\begin{tabular}{lrrrrr}
		\hline
		\textbf{Rates}  & 0.03\% & 0.05\%& 0.1\%& 0.2\% & 0.3\%\\
		\hline
		\makecell{\textbf{GCN}}&\makecell{46.3\\(0.08)}&\makecell{56.1\\(0.10)}&\makecell{63.3\\(0.06)}&\makecell{70.4\\(0.04)}&\makecell{77.1\\(0.02)}\\
		\hline
		\makecell{\textbf{GCN}\\\textbf{SVAT}}&\makecell{52.1\\(0.06)}&\makecell{56.9\\(0.08)}&\makecell{63.5\\(0.07)}&\makecell{71.2\\(0.04)}&\makecell{77.2\\(0.02)}\\
		\hline
		\makecell{\textbf{GCN}\\\textbf{DVAT}}&\makecell{\bf53.3\\(0.06)}&\makecell{\bf58.6\\(0.06)}&\makecell{\bf66.3\\(0.05)}&\makecell{\bf72.2\\(0.03)}&\makecell{\bf77.3\\(0.02)}\\
		\hline
	\end{tabular}
	\caption{Classification Accuracies on PubMed with different label rates. Numbers in bracket are the standard deviation of accuracies.}
	\label{table_weakly_pubmed}
	
\end{table}

More specifically, we list the detailed performances of GCNSVAT and GCNDVAT compared with original GCN under different label rates, which are exhibited in Tables~\ref{table_weakly_cora},~\ref{table_weakly_citeseer} and~\ref{table_weakly_pubmed}, respectively. We report the mean accuracy of 10 runs. The results in tables provide a more sufficient evidence for the effectiveness of our two methods.

\subsection{Effect on Standard Semi-Supervised Learning}

Apart from the experiments under different training sizes, we also test the performance of GCNSVAT and GCNDVAT algorithms in standard semi-supervised setting with standard label rates listed in Table~\ref{table_dataset}. Particularly, we compare our methods with other state-of-the-art methods on the node classification task under standard label rate and the results of baselines are referred from \cite{kipf2016semi}.

\begin{table}[htbp]
	\centering
	\begin{tabular}{lrrrrrr}
		\hline
		\textbf{Method}  & \textbf{CiteSeer} & \textbf{Cora} & \textbf{PubMed} \\
		\hline
		\textbf{ManiReg}       &    60.1&    59.5&    70.7\\
		\textbf{SemiEmb}       &    59.6&    59.0&    71.7\\
		\textbf{LP}            &    45.3&    68.0&    63.0\\
		\textbf{DeepWalk}      &    43.2&    67.2&    65.3\\	
		\textbf{Planetoid}     &    64.7&    75.7&    77.2\\
		\textbf{GCN}           &    68.4&  78.4&    77.3\\
		\hline
		\textbf{GCNSVAT}        &  68.7&   78.5 &    77.5 \\
		\textbf{GCNDVAT}        &\bf{69.3}&\bf{78.6   }&\bf{77.6}\\
		\hline
	\end{tabular}
	\caption{Accuracy under 20 Labels per Class across three datasets.}
	\label{table_standard}
\end{table}

From Table~\ref{table_standard}, it turns out that our GCNDVAT algorithm exhibits the state-of-the-art performance though the improvement are not apparent compared with that in few training sizes, while our GCNSVAT algorithm also shares a similar performance. Through the extensive experiments in semi-supervised learning, we demonstrate thoroughly that VAT suffices to improve the generalization performance of GCNs by additionally providing an adversarial regularization both in semi-supervised setting with few labeled nodes and standard semi-supervised setting.

\subsection{Sensitivity Analysis of Hyper-parameters}

One of the notable advantage of VAT in GCNs is that there are just two scalar-valued hyper-parameters: (1) the perturbation magnitude $\epsilon$ that constraints the norm of adversarial perturbation and (2) the regularization coefficient $\alpha$ that controls the balance between supervised loss $\mathcal{L}_0$ and virtual adversarial loss $\mathbb{E}_{x\sim \mathcal{D}}\mathcal{R}_{\rm vadv}$. We refine the analysis in original VAT~\cite{miyato2018virtual} and theoretically demonstrate the total loss is more sensitive to $\epsilon$ rather than $\alpha$ in the regularization control of GCNs with VAT setting.

Consider the second approximation of virtual adversarial regularization:
\begin{equation} \begin{aligned}
\mathcal{R}_{\rm vadv}(x, \mathcal{D}_l, \mathcal{D}_{ul}, \theta)=&\max_r\{D(r, x, \overline{x}, A;\theta); \|r\|_2 \leq \epsilon\} \\ \approx &\frac{1}{2} \epsilon^2 \lambda_1(x, \overline{x}, A;\theta),
\end{aligned} \end{equation} 
where $\lambda_1(x,\overline{x}, A; \theta)$ is the dominant eigenvalue of Hessian matrix $H(x, \overline{x}, A;\theta)$ of $D$. Substituting this into the objective function, we obtain
\begin{equation} \begin{aligned}
\mathcal{L}_0 +  \alpha \mathbb{E}_{x\sim \mathcal{D}}\mathcal{R}_{\rm vadv}
=&\mathcal{L}_0 + \alpha\frac{1}{N_l+N_{ul}}\sum_{x_* \in \mathcal{D}_l, \mathcal{D}_{ul}}\mathcal{R}_{\rm vadv}(x, \mathcal{D}_l, \mathcal{D}_{ul}, \theta)  \\ 
\approx &\mathcal{L}_0 + \frac{1}{2}\alpha\epsilon^2 \frac{1}{N_l+N_{ul}}\sum_{x_* \in \mathcal{D}_l, \mathcal{D}_{ul}}\lambda_1(x_*, \overline{x}, A;\theta).
\end{aligned} \end{equation}
Thus, the strength of regularization is approximately proportional to $\alpha$ and $\epsilon^2$. In consideration of the regularization term is more sensitive to the change of $\epsilon$, in our experiments we just tune the perturbation $\epsilon$ to control the regularization by fixing $\alpha=1$ for both methods. 

\begin{figure}[htbp]
	\centering
	\centering\includegraphics[width=.8\textwidth,trim=120 20 100 30,clip]{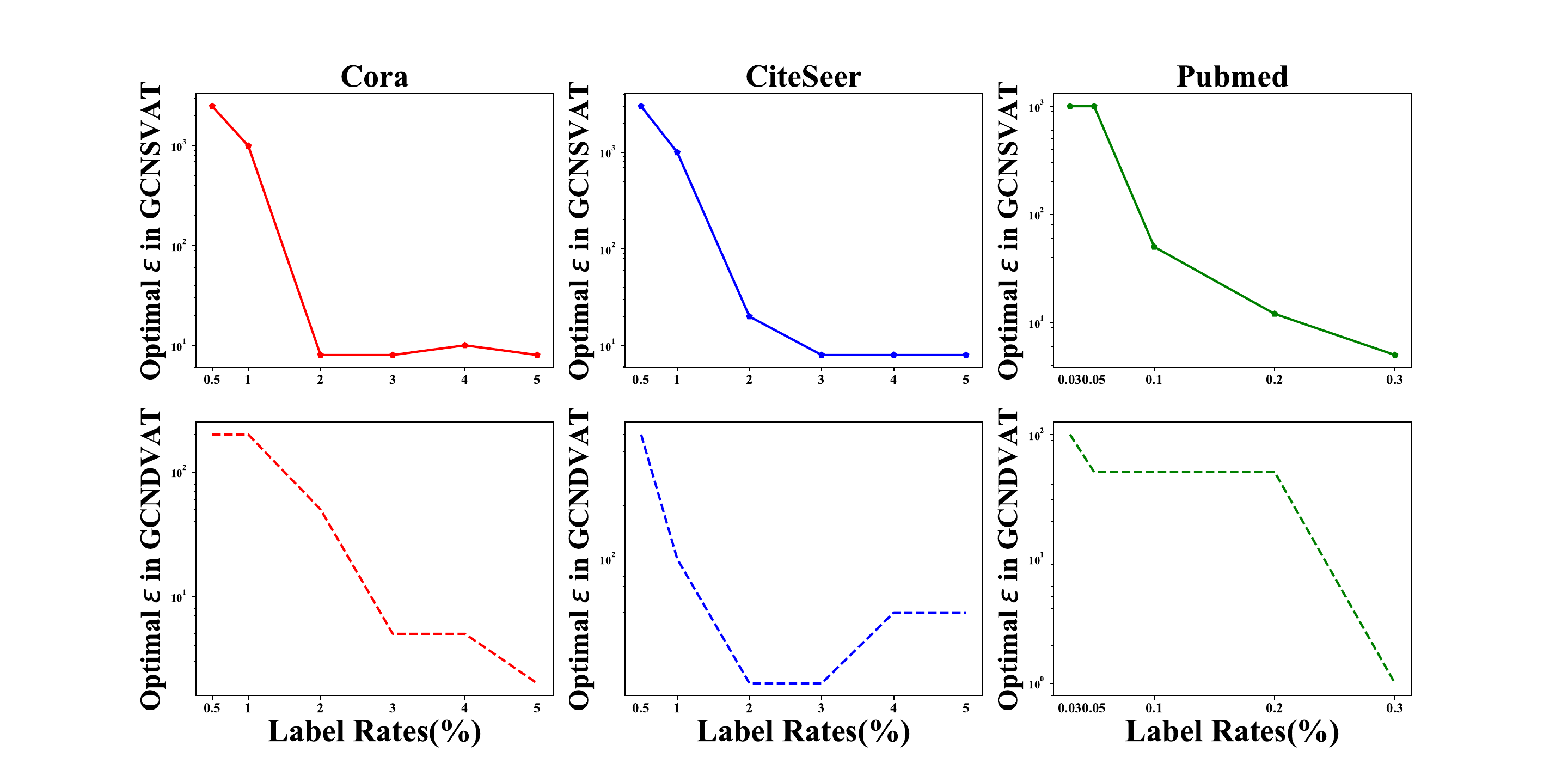}
	\caption{Sensitivity analysis of epsilon $\epsilon$ on two methods.}
	\label{figure_epsilon}
\end{figure}

Further, we present the tendency between the selected optimal $\epsilon$ and label rates. As for the different label rates, it is natural to expect that GCNs with VAT under lower label rate requires larger VAT regularization, yielding the urge for larger optimal $\epsilon$. We empirically verify this conclusion in Figure~\ref{figure_epsilon}.

From Figure~\ref{figure_epsilon}, it is easy to observe that with the increasing of label rates, there is a descending trend of optimal $\epsilon$ for both GCNSVAT and GCNDVAT across three datasets. It meets our expectation since large VAT regularization are more expected for GCNs under lower label rates to obtain the optimal generalization of GCNs. In addition, the optimal $\epsilon$ parameter in GCNSVAT under the same label rate tends to be higher than that in GCNDVAT, especially when the label rate is lower. The reason is obvious because GCNSVAT only applies perturbations on specific elements of sparse feature for each node, thus requiring larger perturbations on those features to get similar regularization effect compared with GCNDVAT.

\section{Discussions and Conclusion}

GCNs with Virtual Adversarial Training is established on the adversarial training on GCNs, which in our paper is simply constrained in the adversarial perturbations on the features of nodes. However, there may exists a better form of adversarial training in GCNs by additionally considering the change of sensitive edges with respects to the output performance. Therefore, incorporating a better form of Virtual Adversarial Training into graphs allows better improvement of generalization of GCNs. Besides, how to combine VAT with other form Graph Neural Networks especially in inductive setting, is also worthwhile to explore in the future.

In our paper, we impose VAT regularization on the original supervised loss of GCN to enhance its generalization in semi-supervised learning, resulting in GCNSVAT and GCNDVAT, whose perturbations are added in sparse and dense features, respectively. Particularly, we apply VAT on GCNs in a theoretical way by additionally imposing virtual adversarial loss on the basic supervised loss of GCNs. Then we empirically demonstrate the improvement caused by the VAT regularization under different training sizes across three datasets. Our endeavour validates that smoothing anisotropic direction on the posterior distribution of GCNs suffices to improve the performance of original GCN model. 

\noindent \textbf{Acknowledgment} ~ Z. Lin is supported by 973 Program of China (grant no. 2015CB352502), NSF of China (grant nos. 61625301 and 61731018), and Beijing Academy of Artificial Intelligence.

\appendix

\bibliographystyle{splncs04} 
\bibliography{samplepaper}
\end{document}